\pdfoutput=1

\documentclass[11pt]{article}

\usepackage[final]{acl}
\usepackage{framed, multido}

\usepackage{times}
\usepackage{latexsym}
\usepackage{graphicx}
\usepackage{array}
\usepackage{multirow}
\usepackage{booktabs}
\usepackage{amsmath}
\usepackage{float}
\usepackage{xcolor}
\usepackage{soul} 
\usepackage{siunitx} 

\definecolor{redtone1}{RGB}{110, 30, 30}
\definecolor{redtone2}{RGB}{188, 136, 136}

\usepackage[T1]{fontenc}

\usepackage[utf8]{inputenc}

\usepackage{microtype}

\usepackage{inconsolata}

\title{Same Task, More Tokens: the Impact of Input Length on\\the Reasoning Performance of Large Language Models}

\author{Mosh Levy\thanks{These authors contributed equally to this work.}$^1$~~Alon Jacoby\footnotemark[1]$^{1}$~~Yoav Goldberg$^{1,2}$\\
$^{1}$Bar-Ilan University\hspace{5mm}
$^{2}$Allen Institute for AI \hspace{5mm}
\\
{\tt \{moshe0110, alonj4\}@gmail.com}}

\begin{document}
\maketitle
\begin{abstract}
This paper explores the impact of extending input lengths on the capabilities of Large Language Models (LLMs).  
Despite LLMs advancements in recent times, their performance consistency across different input lengths is not well understood. 
We investigate this aspect by introducing a novel QA reasoning framework, specifically designed to assess the impact of input length. 
We isolate the effect of input length using multiple versions of the same sample, each being extended with padding of different lengths, types and locations. 
Our findings show a notable degradation in LLMs' reasoning performance at much shorter input lengths than their technical maximum. 
We show that the degradation trend appears in every version of our dataset, although at different intensities.
Additionally, our study reveals that the traditional metric of next word prediction correlates negatively with performance of LLMs' on our reasoning dataset. 
We analyse our results and identify failure modes that can serve as useful guides for future research, potentially informing strategies to address the limitations observed in LLMs.
\end{abstract}

\section{Introduction}
Recent advancements in Large Language Models (LLMs) show impressive performance across a range of tasks \cite{openai2023gpt4,Anil2023GeminiAF,jiang2024mixtral}, including answering correctly complex questions requiring multiple reasoning steps \cite{kojima2022large,wei2022chain}. These models also claim to support increasingly longer inputs.  
This development underscores the need to examine their performance on the longer inputs they are now technically supporting.

A reasonable assumption is that support for long inputs would transfer across tasks and enable a model adept at solving a task when presented in a short input prompt, to perform the same task when it is embedded within a longer prompt. Does this assumption hold?
\begin{figure}[t]
    \centering
    \includegraphics[scale=0.4]{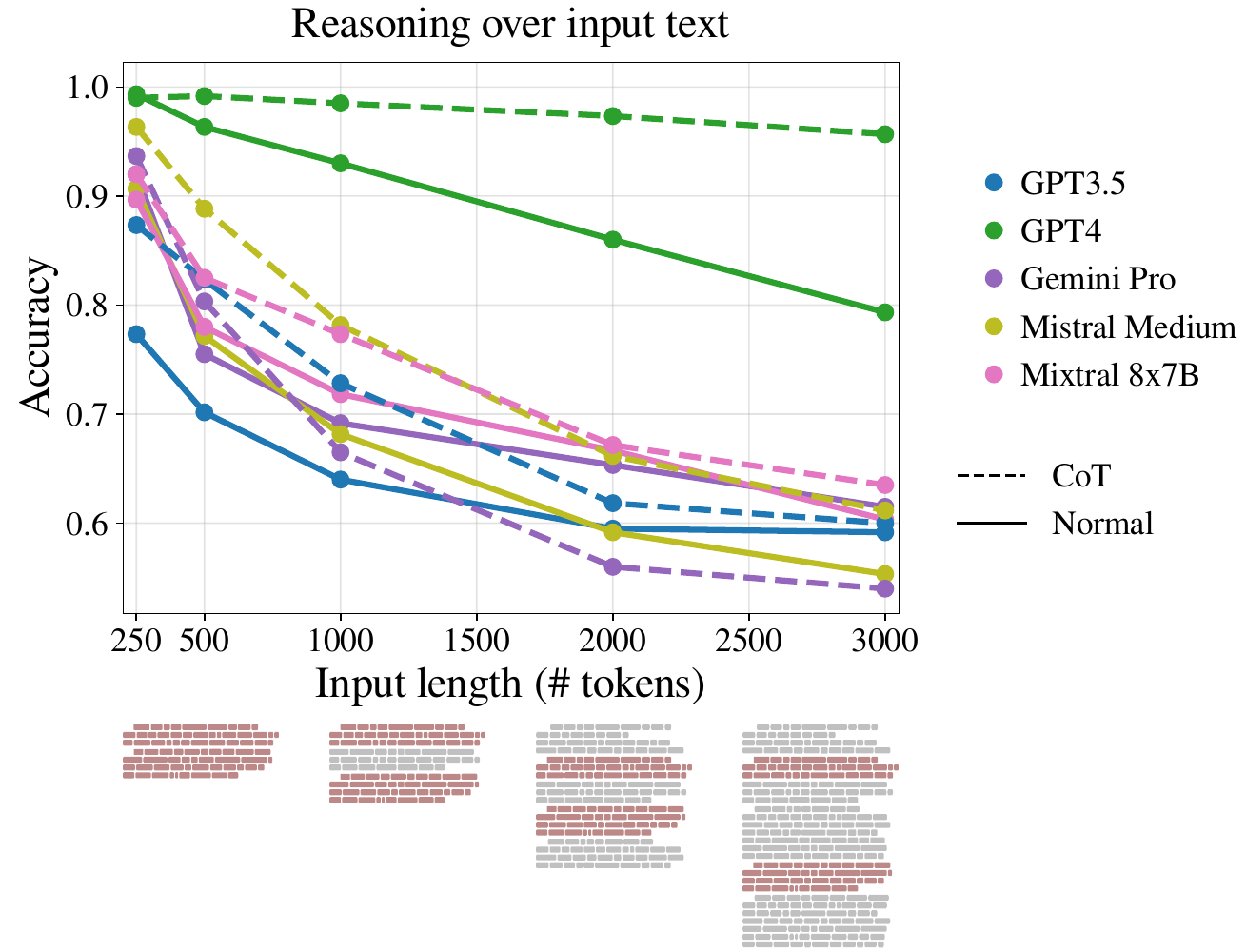}

    \caption{Reasoning performance drops as input grows, across a variety of tasks. Inputs are composed of text containing information relevant to the task (in red), and irrelevant text (grey) which is drawn from various sources and extended incrementally. 
    Two separate text spans are required to answer correctly, and are located randomly in the input.
    Each point reflects the performance across 600 samples.
    }
    \label{fig:introduction_figure}
\end{figure}
Recent studies that benchmark models over tasks that involve longer inputs, including reasoning tasks, indicate that indeed models often struggle with reasoning over long inputs \citep{shaham2023zeroscrolls,Li2023LooGLECL,Bai2023LongBenchAB}.
However, these studies do not properly control their variables, and vary both the input length and the associated tasks to be performed.
This makes it it hard to say if the degraded performance is due to the requirement to work with longer input, or due to the task being generally harder. 

In this work, we study the effect of increasing the input length on model performance, while keeping other factors as constant as possible.

We employ a methodology to measure model performance trends as a function of input length, by isolating it as a variable, while keeping the underlying task intact (\S\ref{sec:desiderata}).

To that end, we introduce \textbf{F}lexible \textbf{LEN}gth \textbf{Q}uestion \textbf{A}nswering dataset (FLenQA) \footnote{\href{https://github.com/alonj/Same-Task-More-Tokens}{https://github.com/alonj/Same-Task-More-Tokens}}, a QA dataset for text-based reasoning (\S\ref{sec:data}). 
For each sample, composed of a True/False question over two pieces of information required to answer it (the context), we create multiple versions of different lengths by embedding the context parts within longer, irrelevant texts.
To ensure that models utilize their entire input, the dataset is composed of tasks for which both pieces of information must reasoned over together in order to correctly answer the question. 
At the same time, we keep the tasks simple enough such that models answer most of them correctly when the information pieces are presented on their own, with no additional padding. 

We show that LLMs quickly degrade in their reasoning capabilities, even on input length of 3000 tokens, which is much shorter than their technical maximum (on average over all tested models, a drop in accuracy from $0.92$ to $0.68$).

Additionally, we explore the effect of embedding the information pieces in various locations within the context, as well as with two kinds of contexts: similar to the information pieces, or dissimilar to them (\S\ref{sec:results}).
We find that regardless of the experimental setting, there are similar trends of degradation.

We also show that next-word prediction performance of models on long inputs is uncorrelated with their performance on downstream tasks of reasoning on long inputs (\S\ref{sec:next_word_prediction}).

Furthermore, we find that while \textit{Chain-of-Thought} (CoT) prompting \citep{kojima2022large,wei2022chain} increases performance in short inputs, in most models it does not mitigate the degradation of performance when inputs are longer: while CoT prompting increases the accuracy over non-CoT prompting, the amount of increase is roughly consistent across context lengths, and is far from closing the performance drop due to long context (\S\ref{sec:chain}). 
The only exception to that is GPT4\footnote{we refer to the models gpt-4-1106-preview, gpt-3.5-turbo-1106 as GPT4 and GPT3.5 accordingly.}, in which the gap between CoT and normal prompting increases as the input is longer.

Finally, we analyse our results and identify several failure modes in model responses (\S\ref{sec:failure_modes}). 
We find that with longer inputs models tend not to follow specific instructions in the input, either providing no answer, or - in the case of CoT prompting - presenting the final answer before outlining the reasoning steps. 
We also observe a bias towards answering "false", as well as a decline in the models' ability to incorporate relevant information in their responses, as input length increases.

\section{Desired Data Properties}
\label{sec:desiderata}

Our goal is to understand how input length affects LLMs reasoning capabilities over text, given that the relevant information remains the same. 
We thus use question answering tasks that require models to reason over a given text.  
For the investigation to be applicable to both open and closed models, we chose a behavioral approach that relies on input intervention \citep{holtzman2023generative}.

We aim for our data to satisfy the following requirements:

\paragraph{Ensuring models reason over the input.}
\label{ensuring}
To examine the performance of models on long inputs, we require that the task can only be solved correctly by drawing conclusions from evidence in the text \citep{huang2022towards}.

\begin{enumerate}
    \item \emph{Each data sample should contain several relevant text spans that are both necessary and sufficient to correctly solve the task}.

    \item \emph{All relevant spans must be consulted jointly to reach a successful solution}. Some tasks, like text summarization, can be solved using a "divide-and-conquer" approach \citep{gidiotis2020divide, liu2022end, wolhandler2022multi}, where each relevant span is individually identified, and then paraphrased and added to the output. We wish to avoid such decomposable tasks, as they do not really require reasoning over long inputs.
    
    \item \emph{The question and supporting relevant spans should consist of novel facts not seen in training.} Ensuring that a task requires reasoning across multiple text spans is a stronger requirement then a task that requires multi hop reasoning. 
    It was shown that models can answer existing reasoning dataset when given one some of the parts that were claimed to be required for the task \citep{chen2019understanding,Min2019CompositionalQD}.
    To avoid model reliance on parametric knowledge when we expect a reasoning process to be done (i.e to avoid data contamination \cite{jacovi-etal-2023-stop,sainz2023nlp}), we desire that an evaluation aimed to test reasoning capabilities will require reasoning over texts that were not previously available.

\end{enumerate}

\paragraph{Isolating the length factor.}
\label{isolating}
To isolate the effect of length, we impose the following requirements:

\begin{enumerate}
    \item \emph{The required reasoning should be independent of the length of the sample}: the relevant spans should remain the same in all length variations.
    \item The \textit{added material} (a.k.a ``padding'', text that is added to control the samples' length) \emph{should not contradict or interfere with the reasoning over the relevant text spans}.
    \item The location of each relevant span within the input should be controllable.
\end{enumerate}

\paragraph{Maintaining natural-looking inputs.}
\label{naturality}
The input should reflect something a user may naturally use in an LLM prompt. For example, a sequence of unrelated sentences is not natural. In contrast, a sequence of unrelated paragraphs but where each paragraph is cohesive is more natural, as such an input may result from collecting relevant information from multiple sources.
To best maintain the naturality of the inputs while changing an input's length, we require that the input should be cohesive at least at the level of paragraphs.

\section{FLenQA}
\label{sec:data}
We introduce the \textbf{F}lexible \textbf{LEN}gth \textbf{Q}uestion \textbf{A}nswering dataset (FLenQA), which follows the requirements set in \S\ref{sec:desiderata}.

FlenQA is composed of three reasoning tasks: Monotone Relations (a new task), People In Rooms (a new task) and a simplified version of Ruletaker \citep{clark2021transformers} (\S\ref{subsec:tasks}).
Each task consists of 100 base instances, from which we create variations of different lengths, different background texts, and different dispersion of facts within the background texts (\S\ref{subsec:variations}).

Each task is completely balanced in its label distribution (``True" and ``False"), and we ensure that most base-instances within it will be solved correctly by the LLMs when presented in their un-expanded forms (\S\ref{minimal_text}).

We release the dataset and the code to generate it from scratch to support future studies of reasoning and long input performance. Generating the dataset from scratch can be used to prevent data contamination in future evaluation. Details and statistics of the tasks appear in Appendix \ref{apx:datasets}.

\subsection{Base instances.}
Each base-instance consists of (1) an \emph{optional prefix} (for example introducing the task or supporting facts); (2) \emph{two key paragraphs}, each of which is thematically coherent and starts with a \emph{key sentence} needed for solving the task; and (3) an \emph{optional suffix} (for example, asking a question about the preceding context).\footnote{The optionality is at the task level, either all instances in the task have a prefix/suffix, or they don't.} For each instance, the different parts are joined by newlines and fed to the LLM.

Throughout the text, key paragraphs are typeset in red, the supporting sentences within them in darker red, and the optional prefixes and suffixes in black.
The full prompts used for each dataset are in Appendix \ref{apx:full_setup}.

\paragraph{Deriving the key paragraphs}
\label{extending_dataset}
Each task relies on two facts, expressed as simple sentences. Each of these sentences is then expanded to a thematically-coherent paragraph, in order to ensure the naturality requirement. This expansion is performed using GPT-4, which we prompt to extend the sentences without adding new information, followed by a manual verification of the results by the authors.

\subsection{The tasks}
\label{subsec:tasks}

\paragraph{Monotone relations (MonoRel)} 
Each key sentence is comparing two person names on monotone scale, e.g. ``X is larger than Y'', ``Y is larger than Z''. The suffix is a True/False question that asks about a relation between two entities that appear in different sentences (they are not explicitly compared in the text). 
The relations are transitive and monotone in nature.
\begin{figure}[H]
    \begin{framed}
        \textbf{MonoRel Example:}\\
        \small
        \texttt{\textcolor{redtone1}{\textbf{Julie Baker is younger than Julian Barton.}} \textcolor{redtone2}{This is a fact that remains constant, unchanging like the northern star. It's a truth that is as clear as day that she} \textbf{...} \\ \textcolor{redtone1}{\textbf{Samantha Arnold is younger than Julie Baker.}} \textcolor{redtone2}{It means that Samantha Arnold has experienced fewer birthdays than Julie Baker. }\textbf{...}}\\
        \texttt{Is Samantha Arnold younger than Julian Barton?}
        \normalsize
    \end{framed}
\end{figure}
This data is inspired by different monotonic relations describing kinship, introduced by \citealt{Sinha2018CompositionalLU}. We define a new set of relation types in this work. 
Following the requirements in \S\ref{ensuring}, answering the question requires reasoning over both key sentences.
The data is created programmatically by randomly drawing names from Faker python library \citep{fakerlibrary} and a relation from a list of hand-crafted relations.

\paragraph{People In Rooms (PIR)}
In each sample in the task, in one key sentence person is said to be located in a named room (``\emph{X is in the old library}"), and the other key sentence describes the room to have a certain property (``the old library has wooden floors''). 
The task is then to infer whether the given person is located in a room with the given property. 

\begin{figure}[H]
    \begin{framed}
        \textbf{PIR Example:}\\
        \small \texttt{\textcolor{redtone1}{\textbf{John's living room is marble-floored}}, \textcolor{redtone2}{a reality that is as intrinsic to the building as its very foundations. The moment } \textbf{...} \\ \textcolor{redtone1}{\textbf{Ethan Washington is in John's living room}}, \textcolor{redtone2}{a fact that has become as much a part of the place as the walls and the ceiling. The truth that Ethan Washington is in John's living }\textbf{...}}\\
        \texttt{Is Ethan Washington in a marble-floored room?}
        \normalsize
    \end{framed}
\end{figure}
This dataset is inspired by the bAbI set of tasks \citep{weston2016towards}, where reasoning is conducted on paths taken by one or more agents. 
PIR is a simplification of the task, involving just one agent.
The names of people in the task are drawn randomly \citep{fakerlibrary}. Rooms and properties were hand selected to be mutually exclusive (for example, a room is either blue-walled or red-walled), so no ambiguous examples are created.

\paragraph{Simplified Ruletaker} 
We employ the task formulation from Ruletaker \cite{clark2021transformers}, a benchmark designed for theorem proving within texts that present explicit logical theories in natural language. Each instance consists of a logical rule, two sentences each introducing a fact, and a question over the rule and facts.\footnote{Initial experiments revealed that most LLMs still struggle with instances involving multiple rules or more than two facts. 
Our Simplified Ruletaker task consists of generated samples that fit these criteria.}

\begin{figure}[H]
    \begin{framed}
        \textbf{Simplified Ruletaker Example:}\\
        \small
        \texttt{Facts:}\\
        \texttt{\textcolor{redtone1}{\textbf{Erin is furry.}} \textcolor{redtone2}{Erin is known for his furriness. He has a lot of fur and  }\textbf{...}}\\
        \texttt{\textcolor{redtone1}{\textbf{Erin is good.}} \textcolor{redtone2}{Erin was always known for how good he is. His goodness appears on all matters of life} \textbf{...}
        }\\
        \texttt{Rule:If X is big and X is good then X is tall.}\\
        \texttt{Question: can the statement "Erin is tall" be derived from the rule and the facts?}
    \normalsize        
    \end{framed}
\end{figure}

\subsection{Length Variations}
\label{subsec:variations}
We expand each base instance to input lengths of roughly 250, 500, 1000, 2000, and 3000 tokens.\footnote{We consider a sample to be of length N if its token count as measured by the GPT4 tokenizer is in $(N-70, N+70)$.} 
To extend the inputs to those targets we add background text that is irrelevant to the question (``padding'', \S\ref{isolating}). For each basic-instance and length pair we create different versions that differ in their source of background text: either \emph{duplicate}, \emph{similar} or \emph{different} than the key paragraphs of the instance. For each of these, we also vary the dispersion of the key-paragraph within the background text.

\subsubsection{Background Texts}

\paragraph{Duplicate.}
\label{duplicate}
To evaluate the extreme case where the length changes but the information remains the same, we perform an experiment where the each length text consists of multiple copies of the key paragraph. We duplicate each key paragraphs without any modification to achieve the target length of the input. The two duplicated paragraphs appear in alternating order until the desired sample length is achieved.
In this case, of the two sub-tasks of QA reasoning - identifying the key information and reasoning over it, the first sub-task is trivial.

\paragraph{Similar: resampling from the same task.}
\label{resampling}
To get background text that is similar to the key paragraphs, we pad using paragraphs sampled from other base instances of the same task.
To avoid creating contradictions, we exclude paragraphs that contain entities appearing in the key paragraphs.
This padding therefore does not produce adversarial or ambiguous versions of the samples. 
This type of padding creates an input that resembles the RAG setup, where the input is composed of independent texts from a similar source \citep{Mao2020GenerationAugmentedRF}.

\paragraph{Different: Book Corpus.}
\label{books}
To get background text that differs from the key paragraphs, we use text from the Books Corpus \citep{bookcorpus}.
We sample a random (continuous) text from the Book Corpus, and inject each of the key paragraphs within it, while respecting sentence boundaries.

\begin{figure}[t]
    \centering
    \includegraphics[scale=0.35]{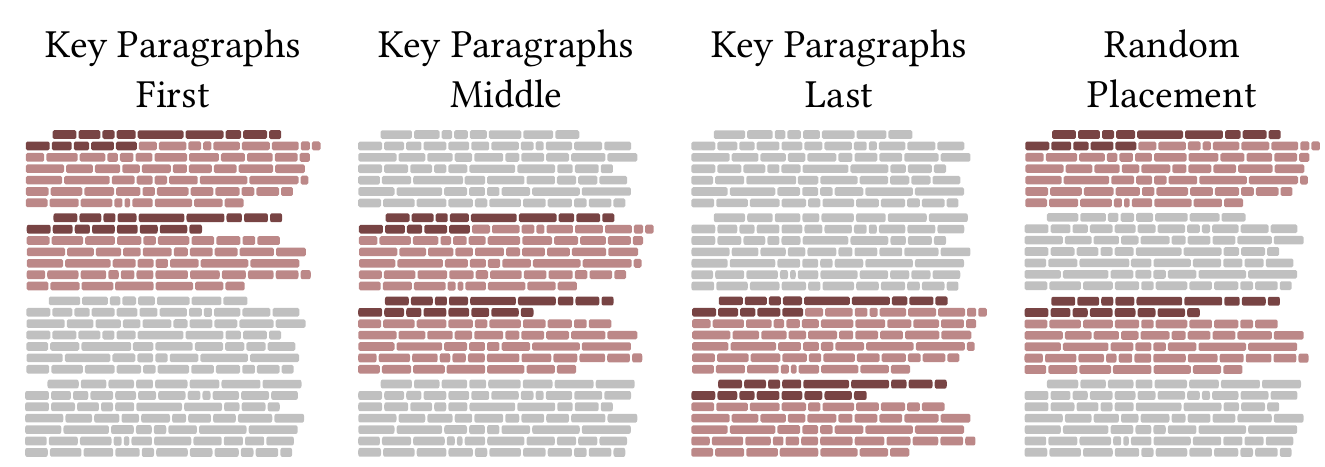}

    \caption{\textbf{Inputs construction.} Key sentences (dark red), are expanded to key paragraphs (light red) which are dispersed in controlled locations among padding text (grey) which is irrelevant to the task.}
    \label{fig:padding}
\end{figure}
\subsubsection{Location of key paragraphs in the text}
\label{padding_dispersion}

We consider four distinct ways in which the key paragraphs are dispersed within the background text: in the first three cases the key paragraphs appear adjacent to each other, while in the fourth the key paragraphs are separated by intervening text of various lengths.

(1) \textit{Key paragraphs first}: The key paragraphs appear at the beginning of the text followed by padding;

(2) \textit{Key paragraphs middle}: Half of the padding is affixed before and half after the key paragraphs, but not between them (the key paragraphs are exactly in the middle);

(3) \textit{Key paragraphs last}: The key paragraphs appear at the end of the text, with padding prepended before them as a prefix;

(4) \textit{Random placement}: padding is added before, between and after the paragraphs, with random intervals.

A visual representation is provided in Figure \ref{fig:padding}.

\subsection{Base instances are answerable}
\label{minimal_text}
We estimate the baseline accuracy by evaluating the LLMs on the minimal text of each sample in the dataset that includes only the question and the key paragraphs relevant to it. The results of the base instances are brought in \ref{tab:minimal}.
We found that even when using non-CoT prompting, four out of the five models achieve high accuracy (>0.89). 
The lowest performing model (GPT3.5) achieve high enough accuracy for degradation to be observable (0.77). 

\label{tab:minimal}

\begin{table}[H]
\centering
\scriptsize 
\begin{tabular}{@{}lccccc@{}}
\toprule
\textbf{Model} & \textbf{Prompt} & \textbf{MonoRel} & \textbf{PIR} & \textbf{Ruletaker*} \\ \midrule

\multirow{2}{*}{GPT3.5} 
& Direct & 0.77 & 0.81 & 0.74 \\
& CoT & 0.86 & 0.88 & 0.88 \\

\multirow{2}{*}{GPT4} 
& Direct & 1.00 & 1.00 & 0.98 \\
& CoT & 1.00 & 1.00 & 0.97 \\

\multirow{2}{*}{Gemini Pro} 
& Direct & 0.84 & 1.00 & 0.92 \\
& CoT & 0.88 & 0.96 & 0.97 \\

\multirow{2}{*}{Mistral Medium} 
& Direct & 0.99 & 1.00 & 0.73 \\
& CoT & 1.00 & 1.00 & 0.89 \\

\multirow{2}{*}{Mixtral 8x7B} 
& Direct & 0.92 & 0.97 & 0.80 \\
& CoT & 0.86 & 0.97 & 0.93 \\

\bottomrule
\end{tabular}
\caption{\textbf{Minimal length accuracy.} The evaluated models have high accuracy on the tasks in our dataset when evaluated on the minimal text (250 tokens). CoT improve performance across almost all tasks and models.}
\label{tab:model_performance}
\end{table}

\section{Main Experiments}
\label{sec:results}
We report average accuracies over all three tasks, and maintain the same setup (prompt, temperature, etc.) over all input lengths. We evaluate five recent capable LLMs: GPT4, GPT3.5, Gemini-Pro, Mistral Medium and Mixtral 8x7B. We consider an output where no answer was mentioned (e.g "I don't know") as incorrect.
See Appendix \ref{apx:full_results} for a detailed breakdown of our setup parameters.

\subsection{Impact of Length and Location}
We start by validating the impact of input length on LLM reasoning performance (Figure \ref{fig:introduction_figure}) in various experimental settings.

\begin{figure}[t]
    \centering
    \includegraphics[scale=0.4]{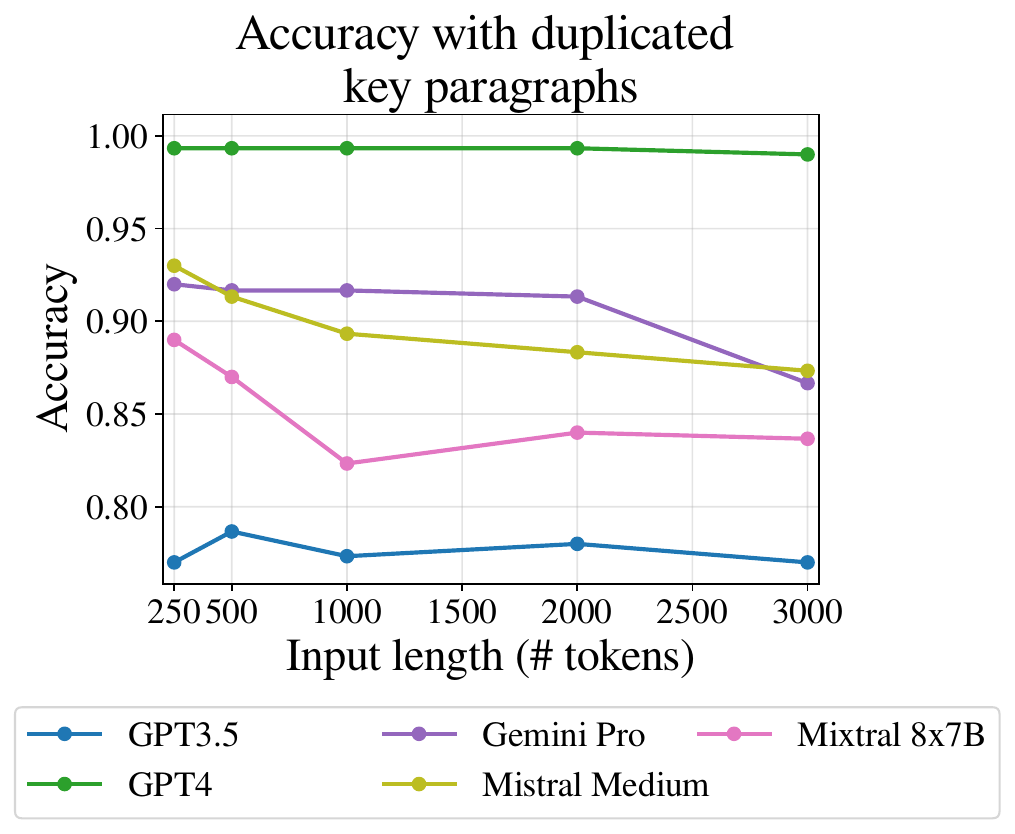}
    \caption{ The relevance of padding is a factor, but it is distinct from the effect of length itself. Some models degrade in reasoning performance. Note, both GPT3.5 and GPT4 are less affected by length when the added tokens are relevant. Each point reflects 300 samples.
}
    \label{fig:identity_control}
\end{figure}

\paragraph{No irrelevant paragraphs}
\label{sub:relevant_tokens}
We first look into the extreme case where only relevant tokens are added (``duplicate padding'').
\citet{shi2023large} Demonstrate that appending irrelevant texts to the input of a reasoning task (GSM-8K \cite{cobbe2021training}) reduces model performance substantially.
We isolate the effect of relevance by testing a setting in which the padding is duplications of the exact text of the key paragraphs.
In this setup, the LLMs are not required to ``search" the input to find the key paragraphs, so any bias towards any position becomes irrelevant \cite{liu2023lost}. Also, any difficulty that might be imposed by the distance between the key paragraphs also becomes irrelevant. 
Hence, we expect that there will be no degradation in performance.
The \emph{Results} shown in Figure \ref{fig:identity_control}, reveal that even in this setup length does play a factor, \emph{and accuracy decreases with length for all models}. 
We consider these results surprising: duplicated texts are an artificial setup which is arguably the best case scenario of long inputs, as the information is constantly repeated and there is no distracting text. 
In more natural cases, most of the input is irrelevant to question asked. We test this setup in the next section.

\paragraph{Adjacent paragraphs surrounded by irrelevant ones} We now move to the more realistic case where the prompt includes the key paragraphs as well as additional irrelevant ones. In the first set of experiments, we keep the key paragraphs adjacent to each other: the LLM just needs to focus and operate on a single area of the input, ignoring the rest.
\citet{liu2023lost} Found that in the task of extractive QA, the position of the answer in the text affects the ability of models to answer correctly.  We thus experiment with the three scenarios: positioning both key paragraphs at the start, end or middle of the text. In all cases we average over both types of irrelevant padding.

\label{sub:pos}
\begin{figure}[h]
    \centering
    \includegraphics[scale=0.4]{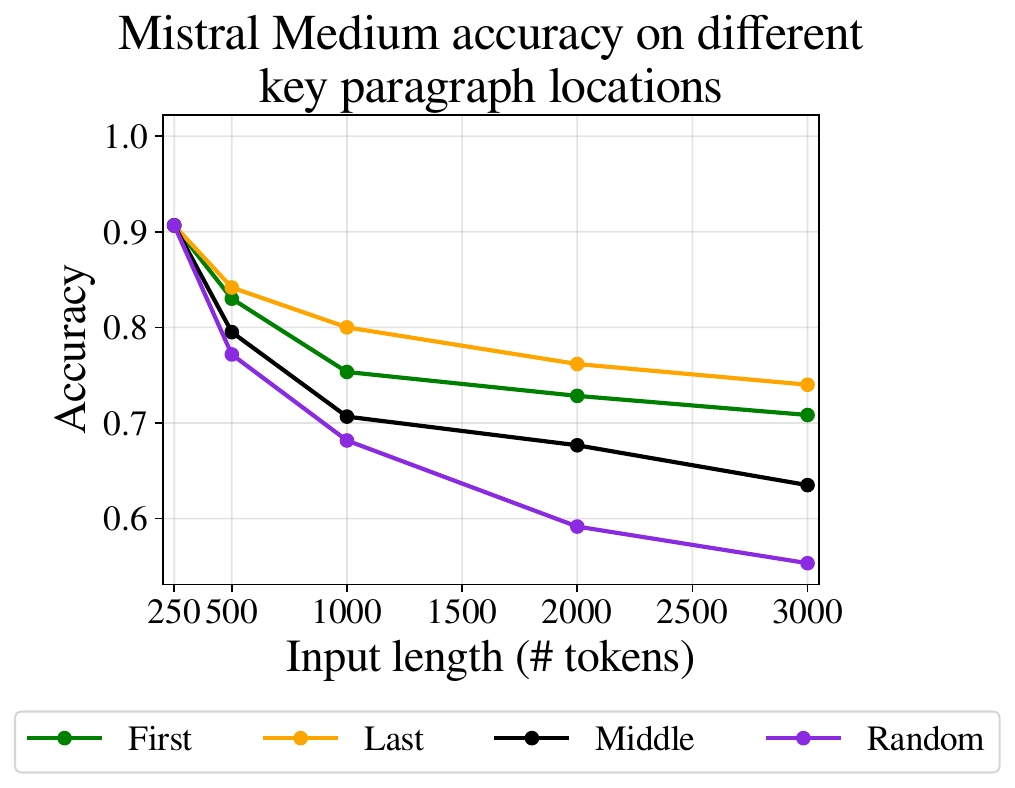}

    \caption{Accuracy decreases as input length grows regardless of where the key paragraphs are placed within the input. Each point reflects 300 samples. Results for all models appear in Appendix \ref{apx:full_results}}
    \label{fig:pos}
\end{figure}

\emph{The results} in Figure \ref{fig:pos} show a significant drop in accuracy as length increase beyond 500 tokens. For most models, adjacency of key paragraphs produces higher accuracy, and when the key paragraphs appear last, accuracy is often highest (suggesting recency bias).
We also find that some models perform worse when the key paragraphs are in the middle, similarly to what was found in the extraction task studied recently \citep{liu2023lost}.

\paragraph{Non-adjacent relevant paragraphs.} Finally, we test the scenario in which the relevant facts needs to be collected from two non-adjacent locations within the text. 

Here, \emph{the results} in Figure \ref{fig:introduction_figure} show a very large drop in performance as length increases, indicating that reasoning tasks becomes significantly harder for LLMs when they need to collect evidence from two distinct locations in a large-ish context length.

\subsection{Kind of irrelevant material}
\label{sub:padding_type}

We now focus only on the non-adjacent key-paragraphs case, and explore the effect of the kind of irrelevant text. 
We consider two scenarios: when the irrelevant paragraphs are \emph{similar} to the relevant ones (taken from the same task), and when they are \emph{different} (taken from the books corpus).

\begin{figure}[h]
    \centering
    \includegraphics[scale=0.4]{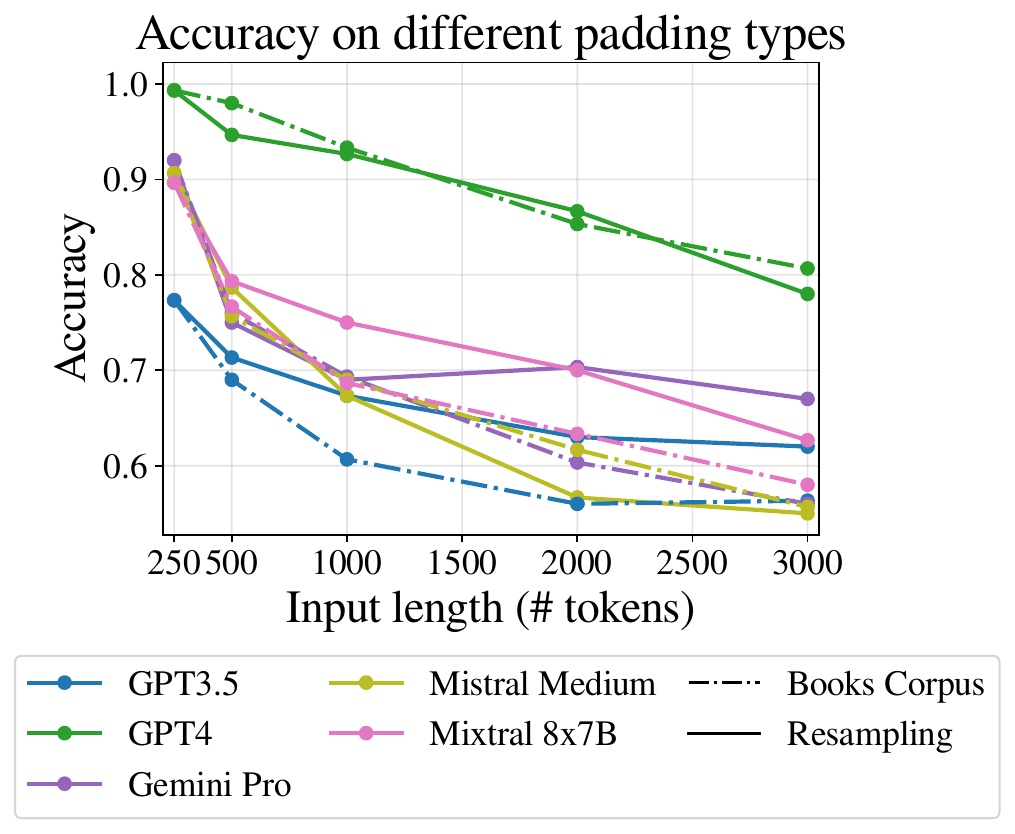}

    \caption{Performance degrade in both types of padding. Books padding impact is much greater in most models. Each point reflects the performance across 300 samples.}
    \label{fig:ILPI_per_padding}
\end{figure}

Our initial expectation was that the setup in which the irrelevant paragraphs are \emph{different} from the relevant ones will be easier for the model, as the irrelevant paragraphs will be easier to discard, aiding focusing on the relevant ones. 
However, the results (Figure \ref{fig:ILPI_per_padding}) show the that is not case: the drop for the \emph{different} setup is mostly larger than for the \emph{similar} one.

\section{Correlation with Next Word Prediction}
\label{sec:next_word_prediction}

Perplexity is used as the main benchmark to show that models utilize their entire input \cite{Anil2023GeminiAF, jiang2024mixtral,ding2024longrope}.
However, it was shown that performance on downstream tasks does not necessarily correlate with model perplexity \citep{liu2023same,xia2022training,tay2022scaling}. 
Here, we will use the flexibility of our dataset to understand the correlation between perplexity and reasoning accuracy.

In closed models we lack access to full vocabulary token probabilities so model perplexity cannot be measured, therefore we resort to measuring next word accuracy on our data. 
We prompt models to complete the next word in a given text, and the output is correct if it is an exact match to the true next word. 
We use the samples in our dataset (without the questions) as the text and compare the results to the reasoning performance on the same samples.

\begin{figure}[h]
    \centering
    \includegraphics[scale=0.4]{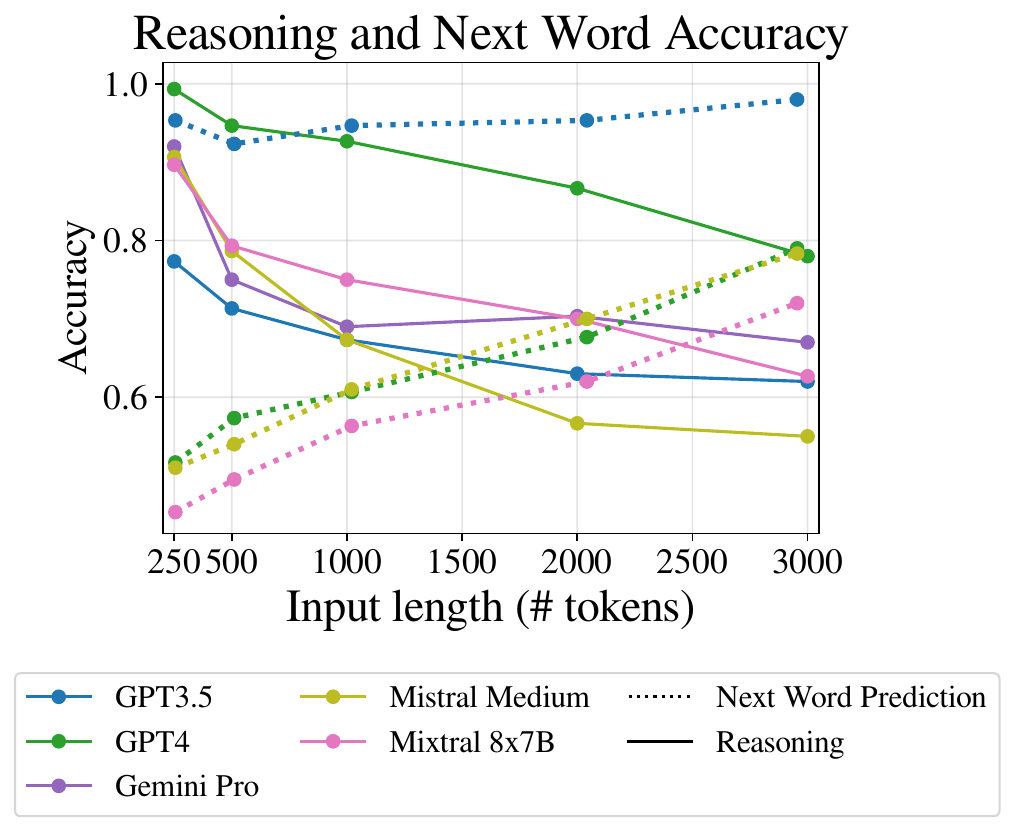}

    \caption{Next word accuracy correlates negatively with the reasoning accuracy on FlenQA. Each point reflects the performance across 300 samples. Gemini-Pro is not included as it answered empty replies to the next word prediction task at any length.}
    \label{fig:NWP}
\end{figure}

Our method finds similar trends on the next word prediction task to those shown in other works \cite{Anil2023GeminiAF, jiang2024mixtral}, namely accuracy increases as input is longer.
However, as shown in Figure \ref{fig:NWP}, next word accuracy correlates negatively with reasoning on FlenQA \footnote{$\rho_{Pearson}=-0.95$, $p=0.01$}.

This implies that measuring next word prediction and, similarly, perplexity, cannot substitute downstream task evaluation on long inputs.

\section{Does Chain of Thought Help?}
\label{sec:chain}
Chain of Thought (CoT) prompting, introduced by \citet{kojima2022large,wei2022chain}, is a technique by which the LLM is pushed to produce a text comprising of reasoning steps before concluding the correct answer for a question. 
\citet{zhou2022large} found that a more specific and optimised instruction ("Let's work this out in a step by step way to be sure we have the right answer.").
 
The CoT technique was shown to significantly improve the accuracy on many reasoning-based question-answering setups. Will using it change the trend and allow the LLMs to perform effectively on longer inputs? We experiment with CoT using the elicitation string of \citet{zhou2022large}.

The results show (Figure \ref{fig:introduction_figure}) that CoT has different effects on different LLMs, and overall does not mitigate the drop in performance due to length. 
In most cases (GPT4, Mixtral 8x7B, Mistral Medium and GPT3.5) it improves performance, but only in GPT4 it has an increased effect as length increases, making it a limited mitigation technique.
In the case of Gemini-Pro, we see that CoT decrease performance as input length is increased, even though it increase performance on short length.

The full results of the CoT prompting over all tasks and setups can be found in Appendix \ref{apx:full_results}.

\section{Length-induced Failure modes}
\label{sec:failure_modes}
We find in the results four \emph{failure modes}:\footnote{All failure modes can be measured automatically using the code in our repository.}
 consistent patterns that correlate with incorrect responses.
 
\paragraph{Failure to answer}
\label{fail_answer}
All of the samples in the dataset can be answered with either "True" or "False", as instructed in our prompts (Appendix \ref{apx:full_setup}).
However, some of LLMs responded with a refusal answer the question, often preceded by a sentence such as "There is not enough information in the text".
\textbf{This tendency grows as the input length increases,} indicating a failure to comply to the instruction that specified a clear choice between "True" and "False".
The trend is demonstrated in figure \ref{fig:false_bias_and_refusal}, and results over all models in Appendix \ref{apx:biases_full_results}.

\begin{figure}[h]
    \centering
    \includegraphics[scale=0.4]{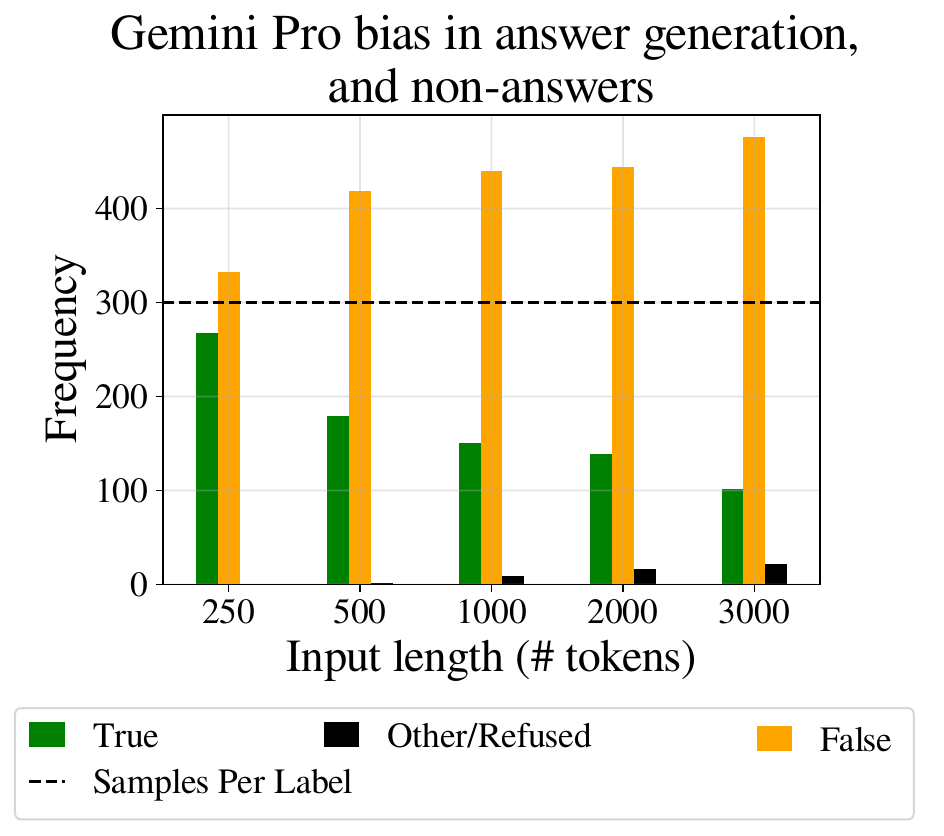}

    \caption{The models exhibit two types of input-length dependent biases: (a) They tend to generate "False" more often than "True", and (b) they ignore instructions and generate answers which do not contain neither.}
    \label{fig:false_bias_and_refusal}
\end{figure}

\paragraph{Label bias}
\label{true_false_bias}
As discussed in \S\ref{sec:data}, our dataset is completely balanced in the label distribution. 
We find that certain LLMs tend to favour one of the labels, typically "false", as the input length grows. 
Results for all models are in Appendix \ref{apx:biases_full_results}.

\paragraph{Answer first, reason later}
\label{answer_first}
When using Chain-of-Thought prompting, some LLMs were much more likely to output the final true/false answer \emph{before} the expected reasoning steps, as inputs grow longer. 
In recent work, \citealt{kojima2022large} found that when models are elicited to provide the reasoning steps after the answer their performance does not increase (as expected when using auto-regressive models that only attend to earlier tokens). 
This can be viewed as a case of failing to follow prompt instructions (see prompt instructions in Appendix \ref{apx:full_setup}) as the length increases. 
In testing, we found that incorrect responses are statistically dependent on the occurrence of answers before the reasoning steps\footnote{Corresponding odds-ratio is 3.643 with $p<0.001$ obtained through Fisher exact test.}.

\begin{figure}[h]
    \centering
    \includegraphics[scale=0.4]{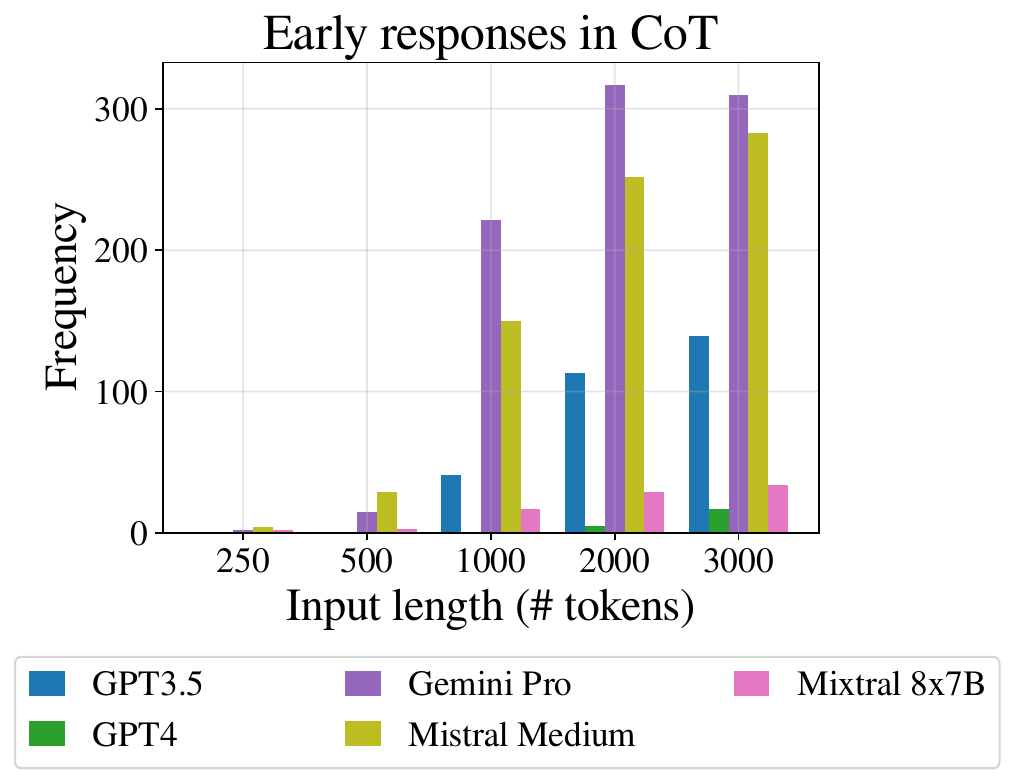}

    \caption{Most of the models tend to generate an answer before the reasoning steps, in a zero-shot CoT prompt setting, and do so more as input length increases.}
    \label{fig:early_response}
\end{figure}

\begin{figure}[h]
    \centering
    \includegraphics[scale=0.4]{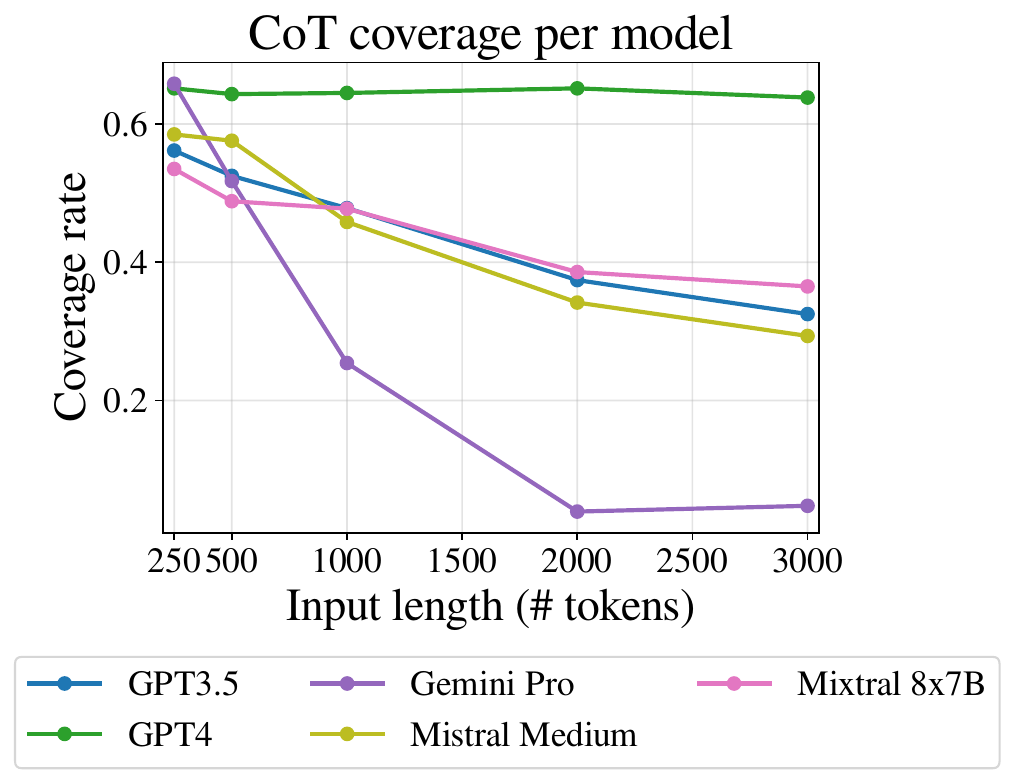}

    \caption{CoT coverage of relevant facts. As input grows, all models fail more often in outputting the task-relevant information at the CoT reasoning steps stage. }
    \label{fig:cot_coverage}
\end{figure}

\paragraph{Chain-of-Thought lack of coverage}
\label{cot_coverage}
All the tasks in FlenQA require the LLM to: (1) locate the relevant texts within the input; and (2) perform the relevant reasoning over them. 
Ideally, the CoT prompt would elicit the LLM to first locate each of the relevant texts and copy them to the ``steps'' part, hence avoiding the effect of long input on reasoning. However, we find that as input length grows, LLMs ability to do this degrades (Figure \ref{fig:cot_coverage}).

We measure this by computing the coverage of the relevant text (the key sentences in each sample) in the models' ``steps" part of the outputs (details in Appendix \ref{apx:coverage}). 
We find that in most models, the ability to locate the relevant text within the input \emph{decreases} as the input length gets longer. 
We found incorrect responses were statistically dependent on the incomplete coverage of the facts\footnote{Corresponding odds-ratio is 3.138 with $p<0.001$ obtained through Fisher exact test.}.

\section{Related Work}
\label{sec:deficiencies}

The evaluation of LLMs on long inputs has followed two distinct pathways: benchmarks of downstream tasks and next word prediction. 
In the realm of benchmarks, studies proposed datasets of long input samples that can be used to evaluate models \cite{shaham2023zeroscrolls,shaham2022scrolls,an2023leval,An2023LEvalIS,Bai2023LongBenchAB}. 
Those datasets are curated over inputs of different, but fixed, length.
This approach, while straightforward, limits our ability to inputs of varying lengths, posing a challenge in understanding the true impact of input length on model performance.
On the other hand, next word prediction evaluations do offer an insights into how models handle inputs of different lengths (like done in \citealt{Anil2023GeminiAF,jiang2024mixtral}). However, the correlation of this task with downstream performance was found not consistent \citep{liu2023same,xia2022training,tay2022scaling}. 
In this paper we reproduce this finding with respect to extended length.

This study builds upon prior research that examined different aspects through input intervention, studying the semantic content (theme) of a task \citep{dasgupta2022language}, prompting strategies \citep{kojima2022large, yao2023tree,jin2024impact} and various properties of the QA task \citep{levy2023guiding}.
Our investigation focuses on input length, isolating it, to reveal its impact on performance.

\section{Discussion}
We study the effect of input length on reasoning performance of current Large Language Models (LLMs).
Our findings reveal a significant drop in performance with longer inputs, occurring well before reaching the models' maximum input-length capacity. 
Our experiments relied on FLenQA, a dataset we constructed that allows to isolate the length factor, by adjusting the parts in the input that are irrelevant to the task. 
We show that regardless of how we adjust the samples, there is still a strong effect of length on reasoning performance.

Finally, we identified specific failure modes, including difficulties in following extended instructions and biases towards less relevant information. 
Our analysis reveals specific failings, providing possible directions for future studies to address and rectify the weaknesses found in LLMs.

In conclusion, our work indicates that evaluating a model's performance based on a single input length does not provide a full picture, and more nuanced evaluation is required. 
We argue that for a model to be considered capable at long range, it must maintain its performance at any length it technically supports.


\section*{Limitations}

Because of the nature of  behavioral testing, the observed drop in performance with varying input lengths remains unexplained; because of lack of access to many of the models, we suspect this direction will continue to be limited. 
Secondly, our approach aimed to create a universally applicable test across different LLMs, leading to the selection of tasks that cater to the lowest common denominator. 
This approach potentially overlooks the nuanced performance differences in more complex reasoning tasks (e.g 5 key paragraphs), where, for instance, stronger models might exhibit performance degradation at shorter input lengths compared to what our findings suggest.
We focused on a subset of reasoning task types which may differ behaviourally from other types. 
Moreover, in order to extend the key sentences to key paragraphs, we employed GPT4 which may introduced some level bias to the text that surrounded the text required to the reasoning task (that was generated without GPT4). 
Finally, our study did not test the distance between key paragraphs, leaving an aspect of LLM performance unexplored that we leave for future research.
\section*{Acknowledgements}
This project has received funding from the European Research Council (ERC) under the European Union's Horizon 2020 research and innovation programme, grant agreement No. 802774 (iEXTRACT).\\
The authors would like to thank Uri Katz, Royi Rassin and Natalie Shapira for illuminating discussions and comments.


\bibliography{custom}

\appendix




We conclude that parametric knowledge should be accounted for when evaluating text-based reasoning capabilities. 
In this work we introduced FlenQA, which is composed of novel generated data to make sure that reasoning over the input is required.

\section{Datasets}
\label{apx:datasets}
Each task of the following contains 100 base instances. 
In each sample, there are two paragraph-length texts (\textit{key paragraphs}). 
To achieve paragraphs of similar length, we edit them by truncating sentences beyond a specific length, resulting in an average paragraph length of 125 tokens.

\subsection{Ruletaker}
The key paragraphs in the task are as evidence for the reasoning task, a rule and a question.
In the original data \cite{clark2021transformers}, the samples contain different number of reasoning steps. 
In this study, we generate new, simpler samples of the task: each sample is composed of only two facts and one logical rule.
The samples we generate are of similar flavor to those that exist in the original Ruletaker data, but are generated with new statements, rules and facts.
The key paragraphs and the padding apear as the facts of each sample.

\begin{figure}[H]
\centering
\small 
\begin{tabular}{@{}ccccc@{}}
\toprule
\textbf{Padding} & \textbf{Target Input} & \textbf{Mean Number} \\
\textbf{Type} & \textbf{Length} & \textbf{Tokens} \\
\midrule

\multirow{5}{*}{Books} & 250 & 249.8 \\
 &  500 & 508.78 \\
 & 1000 & 1009.56 \\
 & 2000 & 2009.64 \\
 & 3000 & 3008.38 \\
\midrule
\multirow{5}{*}{Same} & 250 & 249.8 \\
& 500 & 503.535 \\
& 1000 & 1004.41 \\
& 2000 & 2005.51 \\
& 3000 & 3005.125 \\
\bottomrule
\end{tabular}
\caption{Summary of statistics of the Ruletaker* task data.}
\label{tab:statistics_pir}
\end{figure}

\subsection{MonoRel}
The key paragraphs in the task act as evidence for the reasoning task, and a question. Both key paragraphs describe a monotonic relation between two people, where one person is shared between both. The key paragraphs are embedded in padding text to create a text mixture.

\begin{figure}[H]
\centering
\small 
\begin{tabular}{@{}ccccc@{}}
\toprule
\textbf{Padding}  & \textbf{Target Input} & \textbf{Mean Number} \\
\textbf{Type}  & \textbf{Length} & \textbf{Tokens} \\
\midrule

\multirow{5}{*}{Books} & 250 & 238.06 \\
  & 500 & 490.84 \\
  & 1000 & 991.41 \\
  & 2000 & 1990.34 \\
  & 3000 & 2990.95 \\
\midrule
\multirow{5}{*}{Same} & 250 & 238.06 \\
  & 500 & 491.69 \\
  & 1000 & 991.43 \\
  & 2000 & 1991.31 \\
  & 3000 & 2991.44 \\
\bottomrule
\end{tabular}
\caption{Summary of statistics of the MonoRel task data.}
\label{tab:statistics_MonoRel}
\end{figure}




\subsection{People in Rooms (PIR)}
One key paragraph describes the location of an individual, and the other describes some attribute of that location. The key paragraphs are embedded in padding text to create a text mixture.

\begin{figure}[H]
\centering
\small 
\begin{tabular}{@{}ccccc@{}}
\toprule
\textbf{Padding} & \textbf{Target Input} & \textbf{Mean Number} \\
\textbf{Type} & \textbf{Length} & \textbf{Tokens} \\
\midrule

\multirow{5}{*}{Books} & 250 & 305.36 \\
 &  500 & 491.85 \\
 & 1000 & 989.91 \\
 & 2000 & 1992.00 \\
 & 3000 & 2988.67 \\
\midrule
\multirow{5}{*}{Same} & 250 & 305.36 \\
& 500 & 484.63 \\
& 1000 & 985.82 \\
& 2000 & 1985.04 \\
& 3000 & 2984.80 \\
\bottomrule
\end{tabular}
\caption{Summary of statistics of the People In Rooms (PIR) task data.}
\label{tab:statistics_pir}
\end{figure}


\section{Full Evaluation Setup}
\label{apx:full_setup}
\subsection{Prompts}

\begin{figure}[H]
    \begin{framed}
        \textbf{Ruletaker prompt - Normal:}\\
        \small
        \texttt{Answer whether the statement \{\textit{statement}\} can be derived from the rule and the facts. Answer with either "True" or "False".\\
        Rule: \{\textit{rule}\}\\
        Facts: \{\textit{facts + padding\}\\
        Answer with either "True or "False".}
        }
        \normalsize
    \end{framed}
\end{figure}

\begin{figure}[H]
    \begin{framed}
        \textbf{Ruletaker prompt - CoT:}\\
        \small
        \texttt{Answer whether the statement \{\textit{statement}\} can be derived from the rule and the facts. \\Show your steps then answer with either "True" or "False".\\
        Rule: \{\textit{rule}\}\\
        Facts: \{\textit{facts + padding\}\\
        Answer with either "True or "False". Let's work this out in a step by step way to be sure we have the right answer.}
        }
        \normalsize
    \end{framed}
\end{figure}

\begin{figure}[H]
    \begin{framed}
        \textbf{PIR prompt - Normal:}\\
        \small
        \texttt{
        \{facts + padding\}\\
        True/False Question: \{question\}\\ 
        Answer only True or False.
        }
        \normalsize
    \end{framed}
\end{figure}

\begin{figure}[H]
    \begin{framed}
        \textbf{PIR prompt - CoT:}\\
        \small
        \texttt{
        Show your steps then answer with 'true' or 'false'.\\
        \{facts + padding\}\\
        True/False Question: \{question\}\\ 
        Let's work this out in a step by step way to be sure we have the right answer.
        }
        \normalsize
    \end{framed}
\end{figure}

\begin{figure}[H]
    \begin{framed}
        \textbf{MonoRel prompt - Normal:}\\
        \small
        \texttt{
        Here are some facts. Answer the exact following question based on the text: \{question\} Answer the question as it appears exactly.
        \{facts + padding\}\\
        \{question\}\\ 
        Answer only True or False.
        }
        \normalsize
    \end{framed}
\end{figure}

\begin{figure}[H]
    \begin{framed}
        \textbf{MonoRel prompt - CoT:}\\
        \small
        \texttt{
        Here are some facts. Answer the exact following question based on the text: \{question\} Answer the question as it appears exactly.\\
        Show your steps then answer with 'true' or 'false'.\\
        \{facts + padding\}\\
        \{question\}\\ 
        Let's work this out in a step by step way to be sure we have the right answer. Show your work and finally answer with 'true' or 'false'. The final step should include the exact text of the question and the answer.
        }
        \normalsize
    \end{framed}
\end{figure}

\subsection{Parameters}
All models were evaluated with a temperature of 0 and "top p" of 0 where available to make results as reproducible as possible.
Additionally, We configured Gemini Pro to ignore safety guardrails ("HARM\_CATEGORY" configurations) to overcome its blank answers in some samples.

\subsection{Locating the answer in models' replies}
To identify the models' answers in their responses, we searched for the occurrences of "false" or "true," disregarding case sensitivity. In cases where these words appeared multiple times, only the last instance was considered relevant. We tested the reliability of this method by manually examining a random sample of 100 responses and confirmed its accuracy in all instances.

\subsection{Evaluating the coverage of key facts in CoT}
\label{apx:coverage}
Coverage of the key facts that are relevant to the reasoning task in CoT outputs, was done by searching for (case-insensitive) match of the key sentences in the key paragraphs, within the output of each model. Full coverage means that both key sentences from the input appear in the CoT output. We verified the reliability of this method manually on a sample of 100 responses.

\newpage
\section{Full results}
\label{apx:full_results}
\label{apx:ruletaker_full_results}
\begin{figure}[H]
    \centering
    \includegraphics[scale=0.475]{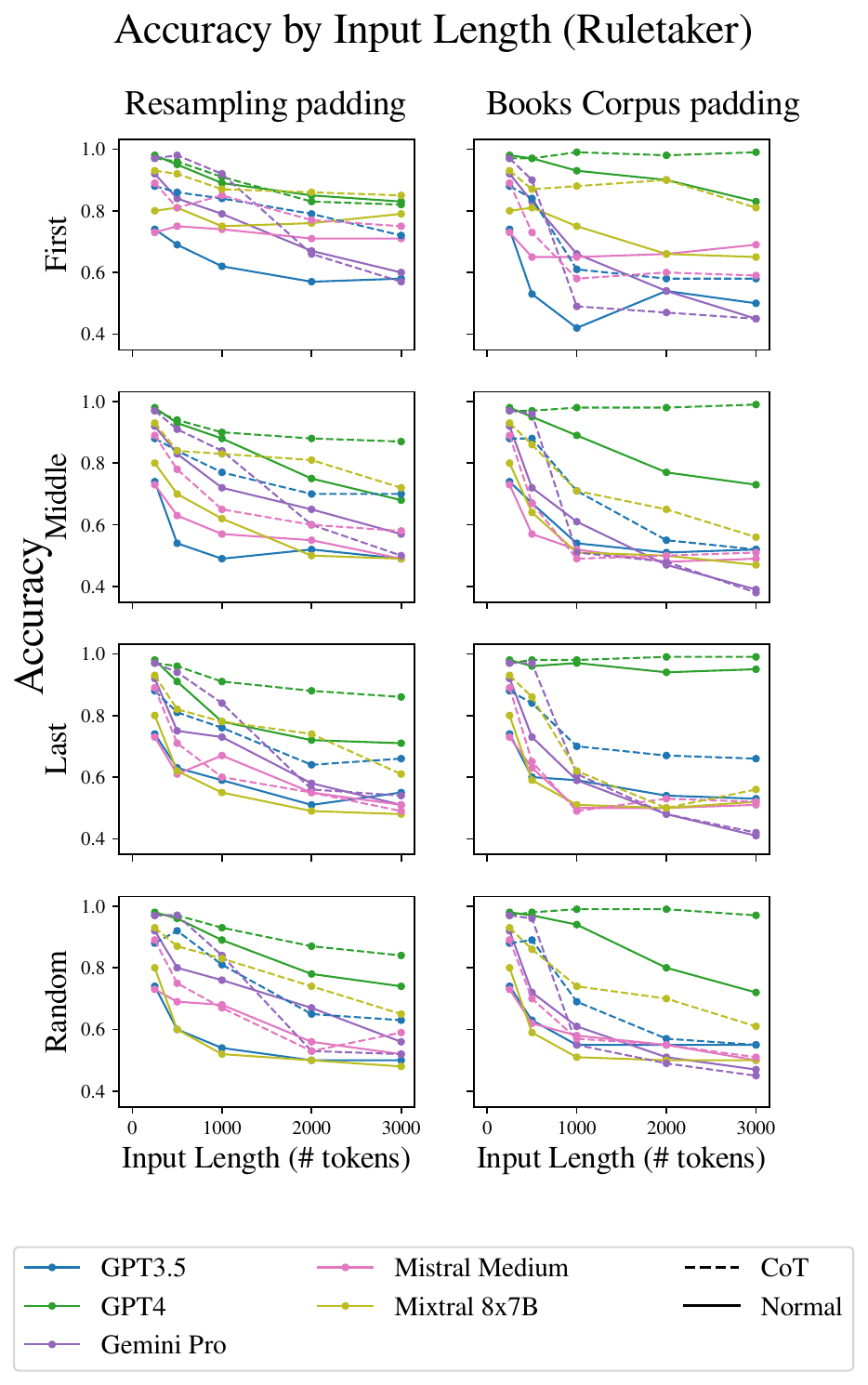}
    \caption{\textbf{Full results for the Ruletaker dataset.}}
    \label{fig:ruletaker_full_results}
\end{figure}

\newpage
\label{apx:monorel_full_results}
\begin{figure}[H]
    \centering
    \includegraphics[scale=0.475]{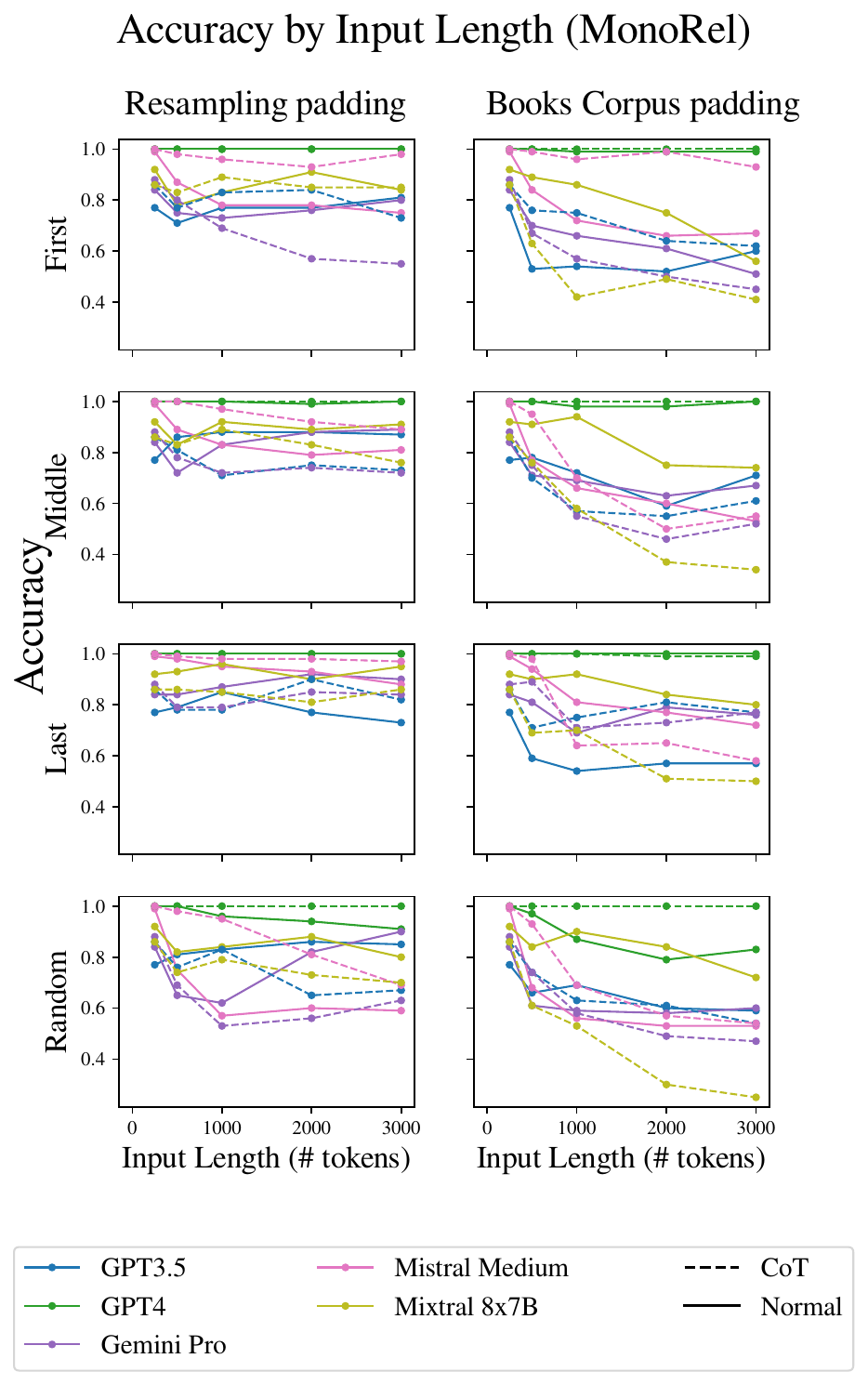}
    \caption{\textbf{Full results for the MonoRel dataset.}}
    \label{fig:monorel_full_results}
\end{figure}

\label{apx:pir_full_results}
\begin{figure}[H]
    \centering
    \includegraphics[scale=0.475]{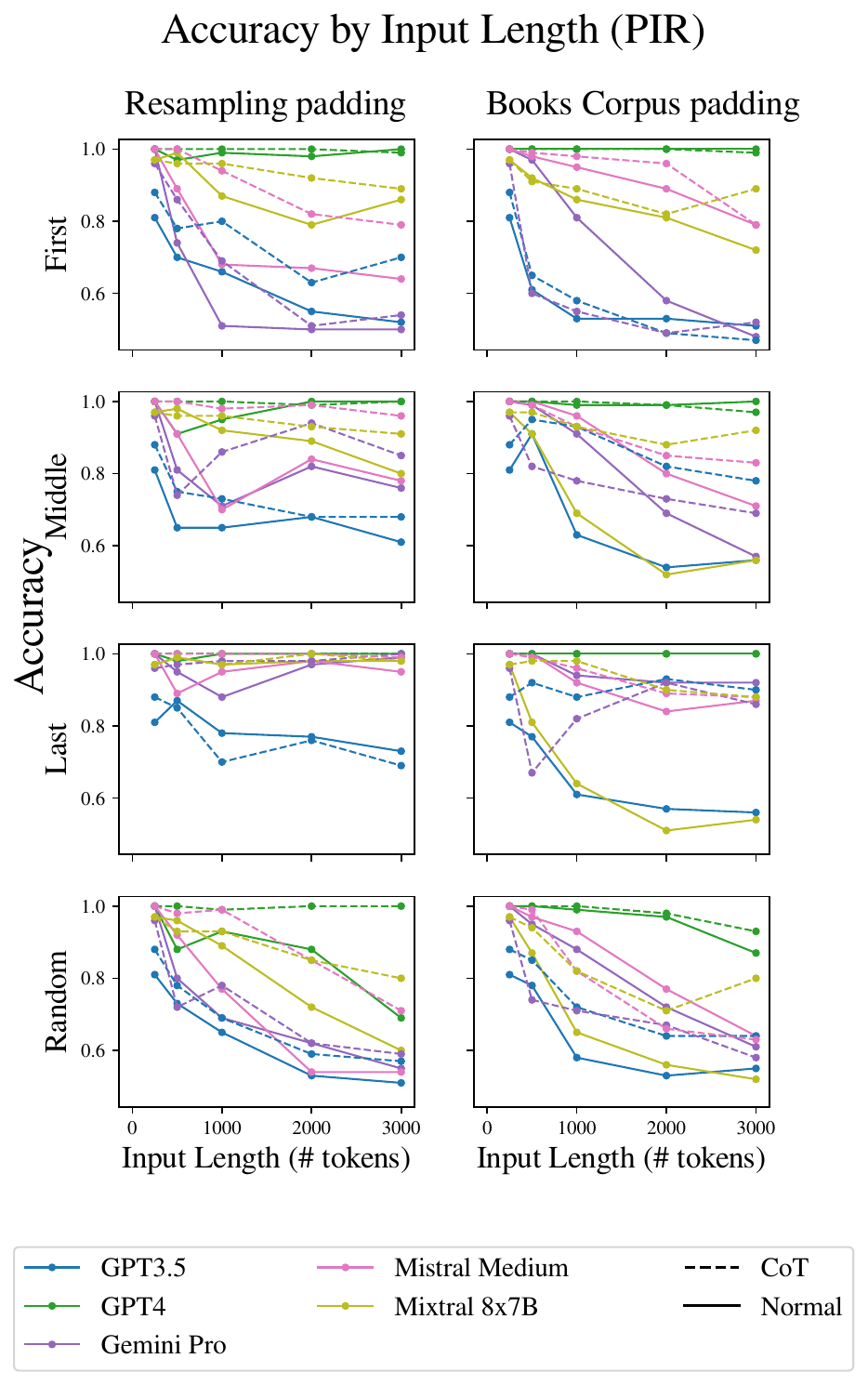}
    \caption{\textbf{Full results for the People In Rooms (PIR) dataset.}}
    \label{fig:pir_full_results}
\end{figure}

\begin{figure}[H]
    \centering
    \includegraphics[scale=0.6]{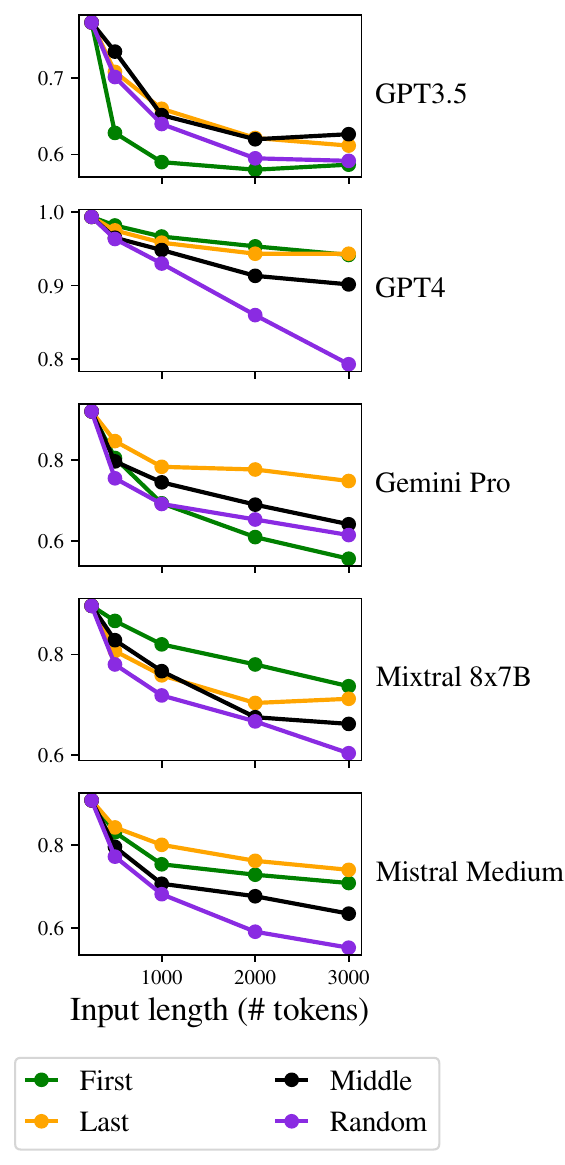}
    \caption{\textbf{Differences in accuracy between different positions of key paragraphs in input.} Averaged over both types of irrelevant padding: similar (resampling from the data) and dissimilar (Books corpus) padding.}
    \label{fig:full_positions}
\end{figure}

\label{apx:biases_full_results}
\begin{figure}[H]
    \centering
    \includegraphics[scale=0.6]{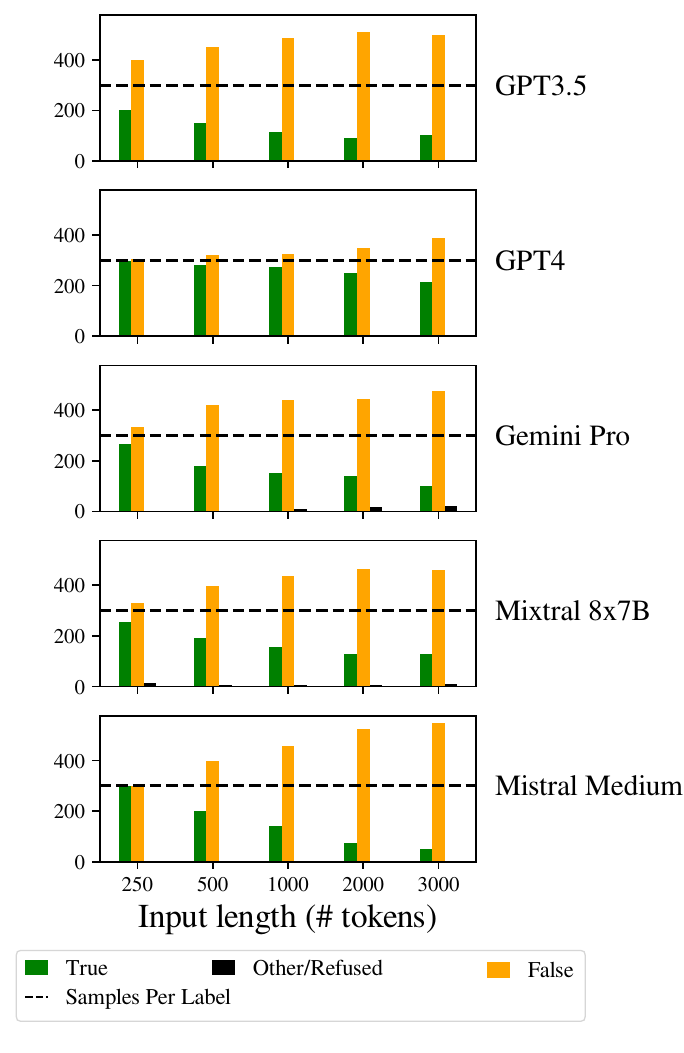}
    \caption{\textbf{Biases in answer generation and non-answers.} Frequency of responses with True, False, or neither, per model.}
    \label{fig:full_biases}
\end{figure}

\end{document}